\documentclass[a4paper,12pt]{article}

\pdfoutput=1

\usepackage[paper=letterpaper,margin=1in]{geometry}

\usepackage{amsmath,amssymb,amsthm,amsfonts,epsfig,cite,setspace,bigstrut,longtable,array,breqn,url,color}

\usepackage{caption,setspace}

\usepackage{tikz}

\usepackage{pdfpages,lipsum}

\begin{document}



\vskip 0.25in

\newcommand{\todo}[1]{{\bf ?????!!!! #1 ?????!!!!}\marginpar{$\Longleftarrow$}}
\newcommand{\sref}[1]{\S~\ref{#1}}
\newcommand{\nn}{\nonumber}
\newcommand{\tr}{\mathop{\rm Tr}}
\newcommand{\comment}[1]{}

\newcommand{\cM}{{\cal M}}
\newcommand{\cW}{{\cal W}}
\newcommand{\cN}{{\cal N}}
\newcommand{\cH}{{\cal H}}
\newcommand{\cK}{{\cal K}}
\newcommand{\cZ}{{\cal Z}}
\newcommand{\cO}{{\cal O}}
\newcommand{\cP}{{\cal P}}
\newcommand{\cR}{{\cal R}}
\newcommand{\cA}{{\cal A}}
\newcommand{\cB}{{\cal B}}
\newcommand{\cC}{{\cal C}}
\newcommand{\cD}{{\cal D}}
\newcommand{\cE}{{\cal E}}
\newcommand{\cF}{{\cal F}}
\newcommand{\cT}{{\cal T}}
\newcommand{\cV}{{\cal V}}
\newcommand{\cX}{{\cal X}}
\newcommand{\IA}{\mathbb{A}}
\newcommand{\IP}{\mathbb{P}}
\newcommand{\IQ}{\mathbb{Q}}
\newcommand{\IH}{\mathbb{H}}
\newcommand{\IK}{\mathbb{K}}
\newcommand{\IR}{\mathbb{R}}
\newcommand{\IC}{\mathbb{C}}
\newcommand{\IF}{\mathbb{F}}
\newcommand{\IV}{\mathbb{V}}
\newcommand{\II}{\mathbb{I}}
\newcommand{\IZ}{\mathbb{Z}}
\newcommand{\re}{{\rm~Re}}
\newcommand{\im}{{\rm~Im}}

\newcommand{\tmat}[1]{{\tiny \left(\begin{matrix} #1 \end{matrix}\right)}}
\newcommand{\mat}[1]{\left(\begin{matrix} #1 \end{matrix}\right)}

\let\oldthebibliography=\thebibliography
\let\endoldthebibliography=\endthebibliography
\renewenvironment{thebibliography}[1]{%
\begin{oldthebibliography}{#1}%
\setlength{\parskip}{0ex}%
\setlength{\itemsep}{0ex}%
}%
{%
\end{oldthebibliography}%
}

\newtheorem{theorem}{\bf THEOREM}
\newtheorem{proposition}{\bf PROPOSITION}
\newtheorem{observation}{\bf OBSERVATION}
\newtheorem{definition}{\bf DEFINITION}
\def\theequation{\thesection.\arabic{equation}}

\newcommand{\setall}{\setcounter{equation}{0}}

\begin{titlepage}

~\\
\vskip 1cm
\begin{center}
{\Large {\bf Machine-Learning Mathematical Structures}}

\medskip

Yang-Hui He

\renewcommand{\arraystretch}{0.5} 
{\small
{\it
\begin{tabular}{rl}
  ${}^{1}$ &
  Merton College, University of Oxford, OX14JD, UK\\
  ${}^{2}$ &
    London Institute of Mathematical Sciences, Royal Institution, London, W1S 4BS, UK\\
  ${}^{3}$ &
    Department of Mathematics, City, University of London, EC1V 0HB, UK\\
  ${}^{4}$ &
    School of Physics, NanKai University, Tianjin, 300071, China \\
    ~\\
\end{tabular}
}}
\url{hey@maths.ox.ac.uk}
\renewcommand{\arraystretch}{1.5} 

\end{center}

\vspace{10mm}

\begin{abstract}
We review, for a general audience, a variety of recent experiments on extracting structure from machine-learning  mathematical data that have been compiled over the years.
Focusing on supervised machine-learning on labeled data from different fields ranging from geometry to representation theory, from combinatorics to number theory, we present a comparative study of the accuracies on different problems.
The paradigm should be useful for conjecture formulation, finding more efficient methods of computation, as well as probing into certain hierarchy of structures in mathematics. 
Based on various colloquia, seminars and conference talks in 2020, this is a contribution to the launch of the journal ``Data Science in the Mathematical Sciences.''
\end{abstract}

\end{titlepage}

\begin{spacing}{0.5}
\tableofcontents
\end{spacing}

\section{Introduction \& Summary}
How does one {\it do} mathematics?
We do not propose this question with the philosophical profundity it deserves, but rather, especially in light of our inexpertise in such foundational issues, merely to draw from some observations on the daily practices of the typical mathematician.
One might first appeal to the seminal programme of Russel and Whitehead \cite{RW} in the axiomatization of mathematics  via systemization of symbolic logic -- perhaps one should go back to Frege's foundations of arithmetic \cite{Frege} or even Leibniz's universal calculus \cite{Leibniz}.
This programme was, however, dealt with a devastating blow by the incompleteness of G\"odel \cite{Godel} and the undecidability of Church-Turing \cite{church,Turing}.

\subsection{Bottom-up}
Most practicing mathematicians, be they geometers or algebraists, are undeterred by the theoretical limits of logic \cite{Minhyong} -- the existence of undecidable statements does not preclude the continued search for the vastness of provable propositions that constitute mathematics.
Indeed, the usage of Turing machines, and now, the computer, to prove theorems, dates to the late 1950s.
The ``Logical Theory Machine''  and ``General Problem Solver'' of Newell-Shaw-Simon \cite{NSS} were able to prove some of the theorems of \cite{RW} and in some sense heralded artificial intelligence (AI) applied to mathematics.

The subsequent development of Type Theory in the 1970s by Martin-L\"of \cite{M-L}, Calculus of Constructions in the 1980s by Coquand \cite{Coquand}, Voevodsky's univalent foundations and homotopy type theory \cite{Voe} in the 2000s, etc., can all go under the rubric of automated theorem proving (ATP), a rich and fruitful subject by itself \cite{Newborn}.
With the dramatic advancement in computational power and AI, modern systems such as Coq \cite{Coq} managed to prove the 4-colour theorem in 2005 and the Feit-Thompson theorem in 2012 \cite{Gonthier}.
Likewise, the Lean system \cite{Lean} has been more recently launched with the intent to gradually march through the basic theorems.
To borrow a term from theoretical physics, one could call all of the above as {\bf bottom-up} mathematics, where one reaches the truisms via constituent logical symbols.

Indeed, irrespective of ATP, the r\^ole of computers in mathematics is of increasing importance.
From the explicit computations which helped resolving the 4-color theorem in 1976 \cite{AHK}, to the completion of the classification of the finite simple groups by Gorenstein et al.~from the mid-1950s to 2004 \cite{Wilson}, to the vast number of software and databases emerging in the last decade or so to aid researchers in geometry \cite{m2,singular,grdb,He:2018jtw}, number theory \cite{magma,lmfdb}, representation theory \cite{gap}, knot theory \cite{knots}, as well as the umbrella MathSage project \cite{sage} etc., it is almost inconceivable that the younger generation of mathematicians would not find the computer as indispensable a tool as pen or chalk.
The ICM panel of 2018 \cite{ICM18} documents a lively and recent discussion on this progress in computer assisted mathematics.

In his 2019 lecture on the ``Future of Mathematics'', Buzzard \cite{buzzard} emphasizes not only the utility, but also the absolute necessity, of using theorem provers, by astutely pointing out two papers from no less than the {\it Annals of Mathematics} which state contradictory results. With the launch of the XenaProject \cite{Xena} using \cite{Lean}, Buzzard and Hale foresee that by the end of this decade, all undergraduate and early PhD level theorems will be formalized and auto-proven.
An even more dramatic view is held by Szegedy \cite{Szegedy} that computers have beaten humans at Chess (1990s), Go (2018), and will beat us in finding and proving new theorems by 2030.

\subsection{Top-Down}
The successes of ATP aside, the biggest critique of using AI to do mathematics is, of course, the current want, if not impossibility, of human ``inspiration''.
Whilst fringing upon so amorphous a concept is both a challenge for the computer and beyond the scope of logic, an inspection of how mathematics {\it has} been done clearly shows that experience and experimentation precedes formalization.
Countless examples come to mind: calculus (C17th) before analysis (C19th), permutations (C19th) before abstract algebra (C19-20th), algebraic geometry (Descartes) before Bourbaki (C20th), etc., \ldots
Even our own presentations in a journal submission are usually not in the order of how the results are obtained in the course of research.

Perhaps ``inspiration'' can be defined as the sum of {\it experience}, {\it experimentation} by trial and error, together with {\it random} firings of thoughts.
Thus phrased, ``inspiration'' perhaps becomes more amenable to the computer.
In this sense, one could think of the brain of Gau\ss\ as the best {\it neural network} of the C19th,  as demonstrated in countless cases. His noticing, at the age of 16 (and based on our definition of inspiration), that $\pi(x) := \#\{p \leq x : p \mbox{ prime} \}$ proceeds approximately as $x / \log(x)$, is an excellent example, whose statement and proof as the prime number theorem (PNT) had to wait for another 50 years when complex analysis became available.
The celebrated conjecture of Birch-Swinnerton-Dyer, one must remember, came about from extensive computer experiments in the 1960s \cite{bsd}; its proof is still perhaps waiting for a new branch of mathematics to be invented.

To borrow again a term in theoretical physics, this approach to mathematics - of gaining insight from a panorama of results and data, combined with inspired experimentation - can be called {\bf top-down}.
One might argue that much of mathematics is done in this fashion.
In this talk \cite{HeTalks}, based on a programme initiated in \cite{He:2017aed}, we will (1) explore how computers can use the recent techniques of data science to aid us with largely top-down mathematics, and (2) speculate on the implications to the bottom-up approach.
To extend the analogy further, one can think of AlphaGo as top-down and AlphaZero, as bottom-up. 

\section*{Acknowledgments}
We are grateful for the kind invitations, in person and over Zoom, of the various institutions over the most extraordinary year of 2020 -- the hospitality and conversations before the lock-down and the opportunity for a glimpse of the outside world during:
Harvard University, Tsinghua University (YCMS, BIMSA), ZheJiang University, Universidad Cat\'{o}lica del Norte Chile, London Institute of Mathematical Sciences, Queen's Belfast, London Triangle@KCL, University of Connecticut, ``Clifford Algebra \& Applications 2020''@UST China, ``String Maths 2020''@Capetown, ``Coral Gables 2020''@Miami, ``International Congress of Mathematical Software 2020''@Braunschweig,  ``East Asia Strings''@Taipei-Seoul-Tokyo, Nankai University, Imperial College London, ``Iberian Strings 2021''@Portugal, and Nottingham University.
We are indebted to STFC UK for grant ST/J00037X/1 and Merton College, Oxford for a quiet corner of paradise.

\section{Mathematical Data}\label{s:data}\setall
In tandem with the formidable projects such as the abovementioned Xena, Coq, or Lean, it is natural to explore the multitude of available mathematical data with the recent advances in ``big data''.
Suppose we were given 100,000 cases of either (a) matrices, or (b) association rules, with a typical example being as follows:
\begin{equation}
(a) 
  {\scriptsize
  \left(\arraycolsep=1pt\def\arraystretch{0.1}
\begin{array}{cccccccccc}
 5 & 3 & 4 & 3 & 5 & 1 & 4 & 4 &
   1 & 2 \\
 5 & 0 & 4 & 5 & 2 & 4 & 4 & 2 &
   2 & 4 \\
 1 & 1 & 2 & 2 & 0 & 4 & 1 & 4 &
   5 & 0 \\
 5 & 0 & 1 & 1 & 0 & 2 & 0 & 5 &
   0 & 1 \\
 2 & 5 & 0 & 1 & 1 & 3 & 2 & 3 &
   0 & 3 \\
 3 & 2 & 2 & 3 & 0 & 0 & 2 & 2 &
   1 & 0 \\
 2 & 2 & 5 & 1 & 4 & 4 & 0 & 0 &
   1 & 2 \\
 5 & 0 & 0 & 0 & 4 & 5 & 0 & 4 &
   1 & 1 \\
 4 & 3 & 4 & 3 & 3 & 1 & 0 & 0 &
   2 & 5 \\
 2 & 0 & 5 & 0 & 3 & 0 & 4 & 4 &
   1 & 5 \\
\end{array}
\right)} \ ,
\quad
(b)
 {\scriptsize
  \left(\arraycolsep=1pt\def\arraystretch{0.1}
\begin{array}{cccccccccc}
 5 & 3 & 4 & 3 & 5 & 1 & 4 & 4 &
   1 & 2 \\
 5 & 0 & 4 & 5 & 2 & 4 & 4 & 2 &
   2 & 4 \\
 1 & 1 & 2 & 2 & 0 & 4 & 1 & 4 &
   5 & 0 \\
 5 & 0 & 1 & 1 & 0 & 2 & 0 & 5 &
   0 & 1 \\
 2 & 5 & 0 & 1 & 1 & 3 & 2 & 3 &
   0 & 3 \\
 3 & 2 & 2 & 3 & 0 & 0 & 2 & 2 &
   1 & 0 \\
 2 & 2 & 5 & 1 & 4 & 4 & 0 & 0 &
   1 & 2 \\
 5 & 0 & 0 & 0 & 4 & 5 & 0 & 4 &
   1 & 1 \\
 4 & 3 & 4 & 3 & 3 & 1 & 0 & 0 &
   2 & 5 \\
 2 & 0 & 5 & 0 & 3 & 0 & 4 & 4 &
   1 & 5 \\
\end{array}
\right)} \longrightarrow 3 \ .
\end{equation}
The matrices could come from any problem, as the adjacency matrix of a directed non-simple graph, or as the map between two terms in a sequence in homology, to name but two.
The association rule could be computing a graph invariant, or the rank of a homology group, respectively.
Such data can then be fed into standard {\bf machine-learning} (ML) algorithms, which excel in finding patterns.
In the parlance of data science, (a) would be called unsupervised ML on unlabeled data and (b), supervised ML, on labeled data.

Having been {\it trained} on large numbers of cases, two questions immediately present themselves:
\begin{description}
\item[Q1: ] {\it Is there a pattern?} This could range from finding clustering to discovering short-cuts to the association rules, all leading to potential conjectures which could then be formulated precisely and hopefully proven. In some sense, this is a top-down question;

\item[Q2: ] {\it Which branch of mathematics is the data likely to have come from?} This bottom-up question could shed some light on the inherent structure of mathematics.
\end{description}
We will present experiments bearing both questions in mind.
This talk will be a status report of the various comparative experiments undertaken in the last couple of years in the aforementioned programme of ML mathematical structure.

\subsection{Methodology}
To be specific, let us comment on the data structure and the method of attack.
First, we will focus on supervised ML (type (b)). One can certainly give the ML free rein to attempt finding patterns with methods such as  dimension reduction, or clustering analysis, which should  be performed on all ensuing examples in future works.
Here, we shall, however, discuss only labeled data.
This is primarily motivated by the speculations on ``experience'' in the introduction.

Extraordinary effort has been engaged, especially over the past 20 years, in creating data-sets in mathematics where requisite quantities have been computed using oftentimes exponential-complexity methods, compiled and made freely available (typically $\sim$ 10 Gb in size and manageable for the contemporary laptop).
This supervision, in telling the ML {\it what} is interesting to calculate (regardless of the {\it how}), imparts the ``experience'' of the mathematical community while leaving the freedom for ``intuition'' to the AI.
After all, is not much of mathematics concerned with how to generate an ``output'' (the label) for an ``input'' (the configuration)?

A great analogy would be the archetypal problem in ML: hand-writing recognition.
Suppose one is given
\begin{equation}
\includegraphics[trim=0mm 0mm 0mm 0mm, clip, width=4in]{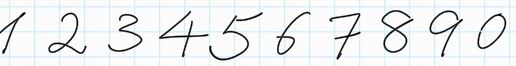}
\end{equation}
and needs to let the computer know that these represent $\{i\}_{i = 0, 1, \ldots, 9}$.
Given these shapes, a mathematician might first think to set up some Morse function to detect critical points, or find a way to calculate some topological invariant. 
This is, of course, highly expensive and also vulnerable to the wide variation in how people write these digits.

How does Google or any modern personal device solve this problem?
Any image is represented by an $n \times n$ matrix each of whose entry is a 3-vector in $[0,1] \times [0,1] \times [0,1]$.
In other words, we have $n^2$ pixels of RGB values.
If one wants only black and white, each entry is simply the gray-scale value in $[0,1]$.
Over the years, long before the recent explosion in AI research \footnote{Incidentally, one of the causes of the recent AI explosion is the success of \cite{Alex} on image recognition which, on utilizing GPUs, has rendered ML efficient to the personal computer.}, NIST (\url{www.nist.gov}) has been collecting handwriting samples (amongst the myriad of other information) by human labeling:
\begin{equation}\label{digits}
\begin{array}{c}
\includegraphics[trim=0mm 0mm 0mm 0mm, clip, width=3in]{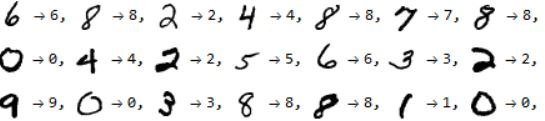} \ldots
\end{array}
\quad
\begin{array}{c}
\fbox{\includegraphics[trim=0mm 0mm 0mm 0mm, clip, width=1in]{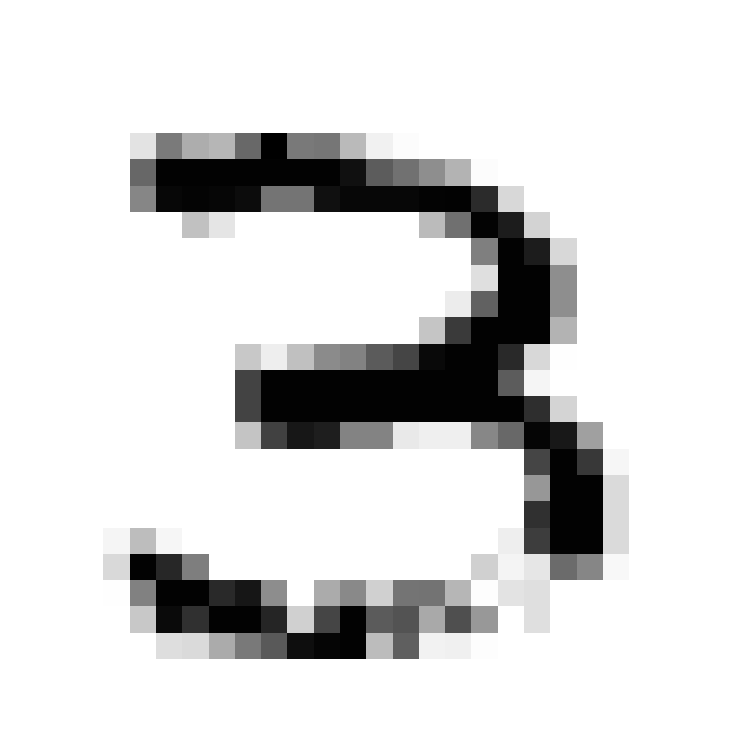}} 28 \times 28 \times (RGB) 
\end{array}
\longrightarrow 3
\end{equation}
The bulk of the complicated task has thus been done \footnote{
When I was a grad student at MIT, I remember Prof. D.~Freedman calling certain problems ``perfectly adapted to large parallel clusters of graduate students.''
} and with only 10000 labeled samples, a standard supervised ML, in this case a convolutional neural network (CNN), could very quickly reach astounding accuracy.
Indeed, the accuracy continue to improve each time a user corrects Google. 

We will not introduce ML here and refer the reader to canons such as \cite{GBA}, and leave a rapid initiation to \cite{He:2020mgx}, as well as longer recent monographs for mathematicians to \cite{He:2018jtw,Ruehle:2020jrk,TTH}. We mention in passing that supervised machine-learning can be thought of as a generalized, non-linear, and not necessarily analytic, regression.
In this sense, Gau{\ss}'s observation on the PNT was an early example of supervised ML.
One remark is that when the output is continuous, an ML algorithm is typically called a {\it regressor}, and when discrete, a {\it classifier}. We will mostly deal with classifiers on discrete/categorical data in our examples below both for uniformity and because regression often requires analytic forms which one might not know a priori \footnote{
There is  a field of symbolic regression which attempts to systematically guess at the functional form, into which we will not delve here.
}.

Next, let us remark on the data.
As well known, the vast majority of the explosion in AI research has been geared toward the human experience, from image processing to medical treatments, from speech recognition to mechanical design, etc. A power of ML, in its {\it emergence of complexity via connectivism} \cite{GBA}, is its ability and effort to deal with ``noise'' and variability, as clearly seen in \eqref{digits}.
The irony is that mathematical data does not quite suffer from this short-coming; there is, by definition and construction, {\it inherent structure} and regularity. A plethora of such data we will shortly encounter and explore.

What is more, outliers are sometimes even more interesting, as exceptional Lie algebras or sporadic groups come to mind \cite{He:2015yoa}.
One constraint we will make, however, is that the range of values in our data, both in the input and the output, be not too great.
Such large variation, especially in the case of integers, as we will see below, tends to make regressors and classifiers struggle.
In principle, we could standardize by only considering binary data with binary labels, and such a study should be undertaken systematically, particularly in light of our forthcoming discussion on hierarchical difficulty.
For now, we will restrict our attention to cases where the entries to our input tensors as well as the output to within the same order of magnitude.

Finally and perhaps most importantly, let us discuss the methodology.
Mathematicians have a ``bag of tricks'' when confronted with a problem and these go under various names.
While results grow steadily throughout history, the fundamental set of ideas and methods increases at a much slower pace.
Hilbert formalized these to a programme of {\it finitary methods}. Landau established the {\it theoretical minimum}. Migdal called them {\it MathMagics}. One can think of these as a standard set of techniques, from analysis to algebra to combinatorics to arithmetic, etc., which, combined together, can tackle the most abstruse of problems.
Again, {\it complexity emerges from inter-connectivity}.
Perhaps hidden in this emergence is the very basis of ``intuition''.

Phrased this way, imitating this set of standard tricks seems natural to ML \footnote{
Interestingly, the IMO Grand Challenge \cite{IMO}, which has just been launched, aims to create an AI algorithm to get a Gold at the International Maths Olympiad.
It was rendered that the IMO presented a perfect set of difficult problems to a limited set of techniques (known to the high school student).
}.
We can therefore take, as our set of methods, some of the standard techniques from supervised ML, to name a few:
\begin{itemize}
\item neural network (NN): we will take care to use only relatively simple architectures such as a feed-forward network (MLP) with only a few layers and only simple activation functions, and without much hyper-parameter tuning.
While a typical such NN is usually represented graphically (with example dimensions to illustrate schematically) as
\[
\begin{array}{c}
\includegraphics[trim=10mm 0mm 0mm 0mm, clip, width=5in]{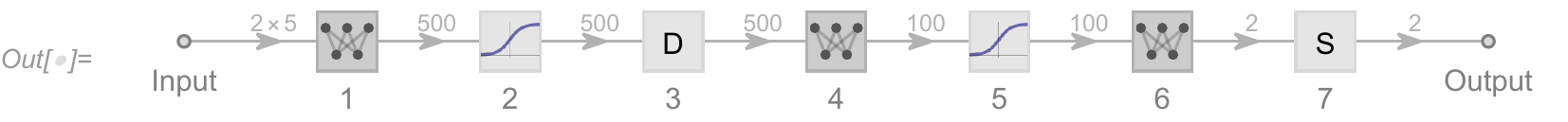}
\end{array}
\]
One can think of this as a composition of maps as
\begin{equation}
I \stackrel{f_0}{\longrightarrow}\mathbb{R}^{n_1}
\stackrel{f_1}{\longrightarrow}
\mathbb{R}^{n_2}\stackrel{f_2}{\longrightarrow} \ldots \mathbb{R}^{n_{k-1}} \stackrel{f_{k-1}}{\longrightarrow}
\mathbb{R}^{n_k} \stackrel{f_k}{\longrightarrow} O 
\end{equation}
with $f_i$ typically as a sigmoid function $\sigma(x)=(1+e^{-x})^{-1}$, or a ReLU function $max(0,x)$.
The integer $k$ is the {\it depth} of the NN and the maximum over $n_i$ is the {\it width}.
The power of ``complexity via connectivism'' can now be very precisely stated in terms of the so-called {\it universal approximation theorems} \cite{UAT} which essentially state that given sufficient depth/width, any input $\to$ output can be approximated to arbitrary precision.

\item support vector machine (SVM): 
this is a very interpretable way to analyze data by finding an optimal hyperplane (and using so-called kernel tricks, hyper-surfaces) which separate data-points with different labels (different categories of configurations).
The hyperplane, whose equation can be written down explicitly, is found by maximizing its distances to points of different labels.

\item Statistical classifier: 
Improving on simple frequency analysis, a remarkably powerful method is that of na\"{\i}ve Bayesian classifiers, where one tracks not only an individual frequency of occurrence, but (assuming independence) also that of sequences in the input collectively.

\item Decision Tree \& Clustering:
One could organize the categorization of the labeled data in a tree-like structure. Similarly, one could find nearest neighbours and clusters in order to classify the input.

\end{itemize}
It is curious that in most of the ensuing experiments, the performance is comparable among most of these above standard methods. That is, the inherent structure of mathematical data responds well to the standard methods.

Performance can be quantified. For discrete output this is usually done as follows.
We have data $\cD = \{ x_I^{(j)} \to d^{(j)} \}$ where $x$ is typically some tensor input with multi-index $I$ and $d$ is the associated output (label); $j$ indexes the data-points.
We split this disjointly into a training set $\cT$ and a validation set $\cV$ so that $\cD = \cT \sqcup \cV$.
Usually, $|\cT|$ is taken \footnote{
A standard thing is to perform 5-fold cross-validation, where the data is randomly divided into 5 equal parts, so that 4 parts can be chosen to be $ \cT$ and the 1 part, $\cV$.
The ML algorithm can then be performed 5 times for the 5 different choices of the 4-part $\cT$, so that an error bar can be collected for the accuracy upon validation.
} to be 80\% of $|\cD|$, and $|\cV|$, the remaining 20\%.
The ML algorithm is applied to $\cT$ (the training of the machine) and then the inputs of $\cV$ are fed so as to give a set of predicted values $\{ \widetilde{d^{(j)}} \}$.
The pairwise comparison between the actual values $d^{(j)}$ and $\widetilde{d^{(j)}}$ for each of the members of $\cV$ is then a measure of how good the ML is on the data.

Since our output $d$ is mostly discrete, say $n$ distinct values (categories), we can write an $n \times n$ matrix with the $(i,j)$-th entry being the number of cases predicted to be $j$ while the actual value is $i$.
This is called a {\it confusion matrix} $M$ which we wish to be as close to diagonal as possible.
One can use na\"{\i}ve precision $p$ (percentage of agreement of $d$ with $\tilde{d}$) in conjunction with confidence (e.g., by the chi-squared $\chi^2$ of $M$, or more precisely, the Matthews' correlation coefficient $\phi := \sqrt{\chi^2/n}$) as a measure of how good the prediction is.
We desire both $p$ and $\phi$ to be as close to 1 as possible. 
Henceforth, we report the pair as a measure of accuracy for all of the experiments, under 80-20 training/validation split:
\begin{equation}
\mbox{Accuracy}   :=  (p, \phi) = \mbox{(na\"{\i}ve precision, Matthews' correlation)} \ .
\end{equation}

\section{Exploring the Landscape of Mathematics}\setall
We have spent too long philosophizing and the advice from Leibniz to Feynman to go and calculate rings in our ears.
The main purpose of this talk is a comparative status report of the results in different branches of mathematical problems since \cite{He:2017aed}.
We will present the precision and confidence of the various experiments while bearing in mind the two questions posed in the beginning of \S\ref{s:data}.

\subsection{Algebraic Geometry over $\IC$}\label{s:geo}
We begin with algebraic geometry over the complex numbers (we emphasize $\IC$ here as we will delve into arithmetic geometry later) for two reasons.
First, dealing with systems of complex multi-variate polynomials is structurally convenient and the algebraic closure of $\IC$ renders such class of problems well behaved in a formal sense.
Second, the initial motivation of \cite{He:2017aed} and, in parallel, the independent works of \cite{Krefl:2017yox,Ruehle:2017mzq,Carifio:2017bov}, was to study the landscape of string theory.
The reason for this is that over the last 30 years or so, theoretical physicists, in alliance with pure and computational mathematicians, have been compiling geometrical quantities inspired by super-string compactification, especially for Calabi-Yau manifolds \cite{He:2018jtw}.
Meanwhile, the combined motivation from the Minimal Model programme and Mirror Symmetry has led algebraic geometers to create large databases of algebraic varieties \cite{CGCR,grdb}.
This is the reason we begin with this seemingly technical subject, which prompted \cite{He:2017aed} to consider supervised ML of mathematics.
The details of the ensuing discussion are not important; and the take-home message is that {\it algebraic varieties} are well-represented by matrices.

\paragraph{Warm-up: } 
We can begin with a baby 0-dimension problem: consider a complex quadratic $a z^2 + b z + c = 0$, for $(a,b,c) \in \IC$.
For simplicity and let us take the coefficients to be Gaussian integers uniformly sampled in the range $\pm10 \pm 10 i$, and check whether there is root multiplicity.
That is, we have labeled data \footnote{
Or, to facilitate an ML which tends to treat real data, we can split the input into real and imaginary parts as
 $\{(\re(a), \im(a), \re(b), \im(b), \re(c), \im(c)) \to r\}$
}
\begin{equation}
\cD = \{(a,b,c) \to r\} \mbox{ with $r = 1$ or $2$ } \ .
\end{equation}
Of course, $r=1$ is much rarer, so we down-sample the number of cases of $r=2$. This technique is called {\bf balancing} and we will make sure all our data are balanced, otherwise there will clearly be prediction bias.
One can readily generate, say $10^6$ cases, remove repeats and down-sample $r=2$ to produce a balanced data $\tilde{\cD}$ of size around 3000 each of $r=1,2$.
At our 80-20 split validation, a decision tree classifier can readily achieve accuracy $\sim (0.98, 0.96)$.
One can be a little more adventurous and demand that $(a,b,c)$ be real and find the number of real roots of the quadratic, in which case a similar level of accuracy is achieved.

\paragraph{Geometric Invariants: } 
To try something more sophisticated we need to appeal to databases of varieties.
As mentioned in the beginning of this section, Calabi-Yau manifolds (CY), or complex, K\"ahler manifolds of zero Ricci curvature, have been a favoured playground \cite{He:2020bfv}.
The simplest CY is the torus, which can algebraically be realized as a cubic in $\IC\IP^2$ -- an elliptic curve.
One can, as the quadratic equation example above, record these as vectors of coefficients; to this we will return in \S\ref{s:arith}.
When computing certain topological invariants, however, it suffices to consider only the multi-degree of the polynomials.
Such representation and the likes thereof, luckily, has been extensively compiled over the decades.

One of the favourite data-bases of Calabi-Yau threefolds are the CICYs, short for Completion Intersection Calabi-Yau manifolds, realized as complex homogeneous multi-degree polynomials in products of complex project spaces.
That is, let the ambient space be $A = \IC\IP^{n_1} \times \ldots \times \IC\IP^{n_m}$, of dimension $n = n_1 +n_2 + \ldots + n_m$ and each having homogeneous coordinates $[x_1^{(r)}:x_2^{(r)}:\ldots:x_{n_r}^{(r)}]$ with the superscript $(r) = n_1, n_2, \ldots, n_m$ indexing the projective space factors.
The Calabi-Yau threefold is then defined as  the complete intersection of $K = n-3$ homogeneous polynomials in the coordinates $x_j^{(r)}$. This information can be succinctly written as
\begin{equation}\label{cicy}
X = 
\left[\begin{array}{c|cccc}
  \IC\IP^{n_1} & q_{1}^{1} & q_{2}^{1} & \ldots & q_{K}^{1} \\
  \IC\IP^{n_2} & q_{1}^{2} & q_{2}^{2} & \ldots & q_{K}^{2} \\
  \vdots & \vdots & \vdots & \ddots & \vdots \\
  \IC\IP^{n_m} & q_{1}^{m} & q_{2}^{m} & \ldots & q_{K}^{m} \\
  \end{array}\right]_{m \times K \ ,}
\quad
\begin{array}{lrl}
\mbox{(i)} && K = \sum\limits_{r=1}^m n_r-3 \ , \\
\mbox{(ii)} && \sum\limits_{j=1}^K q^{r}_{j} = n_r + 1 \ , \ \forall \; r=1, \ldots, m \ ,
\end{array}
\end{equation}
with non-negative integers $q_j^r$.
Condition (i) demands complete intersection, and condition (ii) implies the vanishing of the first Chern class (CY condition), and also renders the column recording the dimensions $n_i$ redundant.
Thus, a homogeneous quintic threefold in $\IC\IP^4$ can be written as $[5]$, the complete intersection of 4 quadrics in $\IC\IP^7$ can be written as $[2,2,2,2]$, etc. Likewise, the cubic elliptic curve can be written as $[3]$.
We remark that the physics, or the Calabi-Yau conditions are not important here: all algebraic varieties can be written in term of such matrices, dropping conditions (i) and (ii).
Furthermore, since we are only keeping track of the degrees, $X$ is really a {\it family} of varieties, as the coefficients of the polynomials define complex structure.

The classification, up to permutation and other geometrical equivalences, of \eqref{cicy} was undertaken in \cite{cicy} in the late 1980s;
they were shown to be finite in number, a total of 7890 configurations, with a maximum of 12 rows, a maximum of 15 columns, and all having entries $q_j^r \in [0,5]$. Interestingly, the best super-computer at the time (at the particle accelerator CERN, to which physicists Candelas et al.~had access) was employed.

A problem of vital interest to both mathematicians and physicists is to compute topological invariants, for which matrix representations like \eqref{cicy} are sufficient (the topological invariant should not depend on mild complex deformations).
For instance \footnote{
Incidentally, the manifold $ \left[{\arraycolsep=1pt\def\arraystretch{0.4}\begin{array}{cc} 1&1 \\ 3&0\\ 0&3\\ \end{array}}\right]$ is the well-known Sch\"on threefold. It is a double elliptic fibration over $\IC\IP^1$ and a self-mirror threefold with Hodge numbers $h^{1,1} = h^{2,1} = 19$. We will return to elliptic fibrations shortly.
}, a typical calculation is that $h^{1,1}\left( \left[{\arraycolsep=1pt\def\arraystretch{0.4}\begin{array}{cc} 1&1 \\ 3&0\\ 0&3\\ \end{array}}\right] \right) = 19$, where $h^{1,1}$ is a Hodge number (a complexified Betti number). 
Indeed, due to index theorems, the Betti numbers are not independent and sum to (with signs) Euler numbers, which are easier to compute. Thus one needs to be judicious in choosing which Hodge numbers to calculate.
Typically, as argued before, we can choose the ones with the least variation in range.

The method to obtain Hodge numbers, and in general rank of cohomology groups, is standard long exact sequence chasing. But this is computationally very expensive. Though most common quantities in algebraic geometry can in principle be obtained from the excellent software such as \cite{m2,singular,sage}, the key component of Gr\"obner basis is a doubly exponential complexity algorithm \footnote{Note that ML techniques are beginning to be used in computing Gr\"obner bases \cite{GBML}.}.
Yet, for various datasets such as the CICYs, the topological quantities have been computed and compiled, using various tricks \cite{cicy}. This is another reason the CICY data-set and those of CY manifolds in general have been gazed upon with renewed zest.

Phrasing the above Hodge computation as the labeled data-point $\left[{\arraycolsep=1pt\def\arraystretch{0.4}\begin{array}{cc} 1&1 \\ 3&0\\ 0&3\\ \end{array}}\right] \to 19$ and recognizing that this is structurally no different from a hand-writing problem of \eqref{digits}, provided the starting point of \cite{He:2017aed}.
Enhancing the data by adding in random row/column permutations, the 8000 or so CICYs can be established into a labeled dataset of the form $\cD = \{M \to h^{1,1}\}$ of size $10^6$, say, where $M$ is the configuration matrix and $h^{1,1}$ is a positive integer ranging from 1 to 19.
To uniformize, we right-bottom pad all configurations with 0 so that all $M$ are $12 \times 15$; giving us a 19-channel classification problem:
\begin{equation}\label{cicyML}
\{
\left[
{\scriptsize
\arraycolsep=1pt\def\arraystretch{0.2}
\begin{array}{ccccccccccccccc}
 1 & 1 & 0 & 0 & 0 & 0 & 0 & 0 & 0 & 0 & 0 & 0 & 0 & 0 &
   0 \\
 0 & 0 & 1 & 0 & 0 & 0 & 1 & 0 & 0 & 0 & 0 & 0 & 0 & 0 &
   0 \\
 1 & 0 & 0 & 1 & 0 & 0 & 0 & 0 & 0 & 0 & 0 & 0 & 0 & 0 &
   0 \\
 0 & 0 & 0 & 0 & 1 & 1 & 0 & 0 & 0 & 0 & 0 & 0 & 0 & 0 &
   0 \\
 0 & 1 & 0 & 1 & 0 & 0 & 0 & 1 & 0 & 0 & 0 & 0 & 0 & 0 &
   0 \\
 1 & 0 & 0 & 0 & 0 & 0 & 0 & 0 & 1 & 1 & 0 & 0 & 0 & 0 &
   0 \\
 0 & 0 & 0 & 0 & 0 & 1 & 0 & 1 & 0 & 0 & 1 & 0 & 0 & 0 &
   0 \\
 0 & 0 & 0 & 0 & 1 & 0 & 1 & 0 & 1 & 0 & 0 & 0 & 0 & 0 &
   0 \\
 0 & 0 & 1 & 0 & 0 & 0 & 0 & 0 & 0 & 1 & 1 & 0 & 0 & 0 &
   0 \\
 0 & 0 & 0 & 0 & 0 & 0 & 0 & 0 & 0 & 0 & 0 & 0 & 0 & 0 &
   0 \\
 0 & 0 & 0 & 0 & 0 & 0 & 0 & 0 & 0 & 0 & 0 & 0 & 0 & 0 &
   0 \\
 0 & 0 & 0 & 0 & 0 & 0 & 0 & 0 & 0 & 0 & 0 & 0 & 0 & 0 &
   0 \\
\end{array}
}
\right] \to 9 \ , \ 
\left[
{\scriptsize
\arraycolsep=1pt\def\arraystretch{0.2}
\begin{array}{ccccccccccccccc}
 1 & 1 & 0 & 0 & 0 & 0 & 0 & 0 & 0 & 0 & 0 & 0 & 0 & 0 &
   0 \\
 3 & 0 & 0 & 0 & 0 & 0 & 0 & 0 & 0 & 0 & 0 & 0 & 0 & 0 &
   0 \\
 0 & 3 & 0 & 0 & 0 & 0 & 0 & 0 & 0 & 0 & 0 & 0 & 0 & 0 &
   0 \\
 0 & 0 & 0 & 0 & 0 & 0 & 0 & 0 & 0 & 0 & 0 & 0 & 0 & 0 &
   0 \\
 0 & 0 & 0 & 0 & 0 & 0 & 0 & 0 & 0 & 0 & 0 & 0 & 0 & 0 &
   0 \\
 0 & 0 & 0 & 0 & 0 & 0 & 0 & 0 & 0 & 0 & 0 & 0 & 0 & 0 &
   0 \\
 0 & 0 & 0 & 0 & 0 & 0 & 0 & 0 & 0 & 0 & 0 & 0 & 0 & 0 &
   0 \\
 0 & 0 & 0 & 0 & 0 & 0 & 0 & 0 & 0 & 0 & 0 & 0 & 0 & 0 &
   0 \\
 0 & 0 & 0 & 0 & 0 & 0 & 0 & 0 & 0 & 0 & 0 & 0 & 0 & 0 &
   0 \\
 0 & 0 & 0 & 0 & 0 & 0 & 0 & 0 & 0 & 0 & 0 & 0 & 0 & 0 &
   0 \\
 0 & 0 & 0 & 0 & 0 & 0 & 0 & 0 & 0 & 0 & 0 & 0 & 0 & 0 &
   0 \\
 0 & 0 & 0 & 0 & 0 & 0 & 0 & 0 & 0 & 0 & 0 & 0 & 0 & 0 &
   0 \\
\end{array}
}
\right] \to 19 \ ,
\ldots
\}
\end{equation}
Relatively simple MLPs and SVMs can perform this task to accuracy $\sim (0.9,0.9)$ \cite{He:2017aed,Bull:2018uow,He:2020lbz}.
Recently, with more sophisticated convolutional NNs, the accuracy has exceed 0.99 \cite{Erbin:2020srm}.
We emphasize again that the computing a topological invariant of any manifold (appropriately embedded as an algebraic variety) can be cast into the form of \eqref{cicyML}. We turned to CY because of their being readily available, similar experiments should be carried out for general problems in geometry.

\paragraph{A Host of Activity: } 
Other ML explorations within the CICYs, such as line-bundle cohomology  \cite{Ruehle:2017mzq,Constantin:2018hvl,Brodie:2019dfx,Larfors:2020ugo,Otsuka:2020nsk,Deen:2020dlf}, distinguishing elliptically fibered manifolds \cite{He:2019vsj}, etc., all of which achieved similar high accuracy and/or improved computation times drastically.
While on the topic of Calabi-Yau manifolds, one cannot resist but mention the tour de force work of Kreuzer-Skarke \cite{Kreuzer:2000xy}, which classified all reflexive polytopes (lattice polytopes with a single interior point such that all facets are distance 1 therefrom) in dimension $n=3$ and 4 up to $SL(n;\IZ)$, generalizing the classical result of the 16 reflexive polygons in $n=2$.
The CY threefold is then realized as a hypersurface (as the canonical divisor) in the toric variety \cite{BB}.
This $n=4$ case is a staggering 473,800,776 in number, which is still being data-mined by many collaborations.
The ML of this set is performed in \cite{Carifio:2017bov,Carifio:2017nyb,Altman:2018zlc,Klaewer:2018sfl}.
Again, the representations of these manifolds are in terms of integer matrices: the vector coordinates of the vertices of the polytopes.

Likewise, explorations in Calabi-Yau volumes \cite{Krefl:2017yox}, 
numerical K\"ahler metrics \cite{Ashmore:2019wzb,Anderson:2020hux,Douglas:2020hpv,Jejjala:2020wcc},
Jones polynomials and hyperbolic volumes \cite{Craven:2020bdz} as well as knot invariants \cite{Gukov:2020qaj}, have all met with admirable success.
In summary, this class of problems, involving algebraic varieties, their topological invariants, metrics and volumes, tends to produce data that is well adapted to our ML paradigm.
It is therefore fortunate that such class of problems is a favourite for theoretical physicists, especially string theorists, as algebraic and differential geometry are undoubtedly the correct language to describe Nature: general relativity being a manifestation of Riemannian geometry and elementary particle physics, of gauge connections and bundles (cf.~a recent attempt in summarizing this dialogue \cite{Yang:2019kup}).
There have of late also been daring and intriguing proposals that quantum field theories and space-time itself, are NNs \cite{QFTNN}.

\subsection{Representation Theory}\label{s:rep}
\paragraph{Warm-up: } 
With geometry behaving well to our pattern search, it is natural to wonder about algebra.
Again, we begin with a baby example. Let us take even versus odd functions.
Let $(x,y)$ be random real pairs uniformed distributed over a rectangle, say $[0, \pi] \times [-1, 1]$.
Consider the 4-vectors $(x,y, -x, y)$ and $(x,y, -x, -y)$; the former models an even function, and the latter, odd.
We thus have a dataset, say ranomly sampled to size $10^5$, of points in $\IR^4$, labeled into 2 categories:
\begin{equation}
\cD = \{ (x,y, -x, y) \to 1 \, (x,y, -x, -y) \to 0 \} \ , \quad (x,y) \in [0, \pi] \times [-1, 1] \ .
\end{equation}
A simple SVM can readily achieve accuracy \footnote{\label{ft:modp}
Let us consider another representation for even/odd.
Suppose we {\it fix} a number $p$ and establish a labeled data-set $\{n_i\} \rightarrow n \bmod p $, where $n_i$ is the list of digits of $n$ in some base (it turns out that which base is not important).
A simple classifier such as logistic regression, or an MLP with a linear layer and a sigmoid layer, will very quickly ``learn'' this to accuracy and confidence very close to 1 for $p=2$ (even/odd). The higher the $p$, the more the categories to classify, and the accuracy decrease as expected.
However, if we do not fix $p$ and feed the classifier with pairs $(n,p)$ all mixed up \cite{Minhyong}, then, the accuracy is nearly zero.
That arthmetic properties are hard to ML will be the subject of \S\ref{s:arith}.
} exceeding $(0.99,0.99)$.
This, and more sophisticated symmetry detection in various contexts were done \cite{Chen:2020dxg,Krippendorf:2020gny}.

\paragraph{Finite Groups: } 
From symmetries, we obviously proceed to finite groups (and finite rings) \cite{He:2019nzx}.
Now, a finite group of size $n$ is defined by its $n \times n$ Cayley multiplication table, which is a Latin square (Sudoku solution), i.e., each row and column is a permutation $1, 2, \ldots, n$ with each number appearing exactly once.
However, not every Latin square is a Cayley table of a finite group -- we must have associativity built in.
Of course, there are standard theorems to check this but we will not do so.
For uniformity, let us consider, say, $n=12$;  there are 5 non-isomorphic groups of this size.

We can thus generate a data-set as follows:
consider $12 \times 12$ Latin squares which are not the Cayley tables of any of the groups, perform random permutations on the row and columns independently; label all these with 0;
likewise, consider those which {\it are} truly the groups, perform random permutations, and label these matrices with 1.
Again, the non-Cayley tables vastly dominate so we need to perform less permutations for these to produce a balanced set.
An SVM classifier, for instance, can distinguish the 0 and 1 cases with accuracy $\sim (0.96, 0.92)$.

Perhaps a more striking case study is that of {\bf finite simple groups}, to which we alluded in the introduction.
It would be understandably fascinating if some ML algorithm can tell a simple group from a non-simple one.
Ordinarily, one would have to go through Sylow theorems or compute the character table.
Here, let us see if ML can do so by ``looking'' at the Cayley table.

A preliminary study was initiated in \cite{He:2019nzx} by taking all finite groups up to size 70, say, compute all their Cayley tables using \cite{gap}, and then enhance with random row and column permutations.
Note that up to size 70, there are 602 groups, but only 20 are simple, thus we need to balance the data by permuting the tables for simple groups more. One can easily establish around $10^5$ matrices this way, approximately evenly divided amongst the simple and non-simples.
For uniformity, we bottom-right pad the results with 0 so that we have a binary classification (1 for simple versus 0 for non-simple) problem of $70 \times 70$ integer matrices (Latin squares).
To give a few examples, we have 
\begin{equation}
\{
\left(\arraycolsep=1pt\def\arraystretch{0.2}
\begin{array}{cccc|l}
 1 & 2 & 3 & 4 & 0 \\
 2 & 1 & 4 & 3 & 0 \\
 3 & 4 & 1 & 2 & 0 \\
 4 & 3 & 2 & 1 & 0 \\ \hline
 0 & 0 & 0 & 0 & \mbox{{\Huge 0}}_{66 \times 66} \\ 
\end{array}
\right) \to 0 
\ ,
\left(\arraycolsep=1pt\def\arraystretch{0.2}
\begin{array}{ccccc|l}
 1 & 2 & 3 & 4 & 5 & 0 \\
 2 & 3 & 4 & 5 & 1 & 0 \\
 3 & 4 & 5 & 1 & 2 & 0 \\
 4 & 5 & 1 & 2 & 3 & 0 \\
 5 & 1 & 2 & 3 & 4 & 0 \\ \hline
 0 & 0 & 0 & 0 & 0 & \mbox{{\Huge 0}}_{65 \times 65} \\ 
\end{array}
\right) \to 1
\ ,
\left(\arraycolsep=1pt\def\arraystretch{0.2}
\begin{array}{cccccccc|l}
 1 & 2 & 3 & 4 & 5 & 6 & 7 & 8 & 0 \\
 2 & 4 & 5 & 6 & 7 & 1 & 8 & 3 & 0 \\
 3 & 8 & 4 & 7 & 2 & 5 & 1 & 6 & 0 \\
 4 & 6 & 7 & 1 & 8 & 2 & 3 & 5 & 0 \\
 5 & 3 & 6 & 8 & 4 & 7 & 2 & 1 & 0 \\
 6 & 1 & 8 & 2 & 3 & 4 & 5 & 7 & 0 \\
 7 & 5 & 1 & 3 & 6 & 8 & 4 & 2 & 0 \\
 8 & 7 & 2 & 5 & 1 & 3 & 6 & 4 & 0 \\ \hline
 0 & 0 & 0 & 0 & 0 & 0 & 0 & 0 & \mbox{{\Huge 0}}_{62 \times 62} \\
\end{array}
\right) \to 0
\ , \ldots
\}
\ ,
\end{equation}
corresponding, respectively, to the Klein viergruppe $C_2 \times C_2$, the cyclic group $C_5$ , the quaternion group of size 8, etc. 
Surprisingly, in a matter of minutes on an ordinary laptop, an SVM (with a Gaussian kernel) could perform this task to accuracy $\gtrapprox (0.98, 0.96)$.
Whilst we need to include more groups in this study (which, sadly becomes computationally and memory intensive, since Cayley tables grow as $n^2$), but the investigations hint at the remarkable possibility that
\begin{quote}
{\bf Proto-conjecture: }
{\it 
Consider the (infinite dimensional) space of finite groups, represented appropriately (e.g., by having the Cayley table flattened to vectors in $\IZ^{n^2}$), then there is hyper-surface separating the simple groups from the non-simple groups.
}
\end{quote}
Fixing an $n$, one can consider all groups of order less than $n$, mix them up and balance the data as discussed; the explicit hyper-surfaces have been computed \cite{He:2019nzx} and work is in progress to understand them.

\paragraph{Continuous Groups: } 
What about continuous groups?
Experiments inspired by standard computations in Lie groups were undertaken in \cite{Chen:2020jjw} (to be concrete, classical groups of type $ABCD$, as well as the exceptional group $G_2$ were explored).
Two of the most important calculations in representation theory, especially that of Lie groups (and especially for mathematical physics), are (1) branching rules: the decomposition of a representation $R$ for a group $G$ to that of its maximal subgroup $H$; and (2) tensor products: given a group $G$ and two of its representations $R_{1,2}$, decompose $R_1 \otimes R_2$ into irreps.
Again, there is a convenient way to encode this data: every representation $R$ of a Lie group is uniquely written in terms of a weight vector $v_R \in \IZ_{\geq 0}^r$ where $r$ is the rank of the group.
In this way, a typical example of calculation (2) would be as follows.
Take $G = A_2 = SU(3)$, the tensor decomposition ${\bf 3} \otimes {\bf 15} = {\bf 8} \oplus {\bf 10} \oplus {\bf 27}$ can be phrased as
\begin{equation}
\left( [0,1] \ , [2,1] \right) \longrightarrow \left( [1,1] \ , [0, 3] \ , [2,2] \right) \ .
\end{equation}
Such data can be readily obtained from \cite{LieArt}, even though the computation time is exponential against dimension of the representation.
While it might be difficult to obtain the precise decomposition due to large variation in output (something which would be interesting to investigate), predicting numerical quantities such as the number of terms in the decomposition, for calculations of both types (1) and (2), were found  to be efficient and accuracy $\sim (0.96, 0.9)$ can be achieved \cite{Chen:2020jjw}.

\subsection{Combinatorics}
From geometry and algebra, we move on to combinatorics and graph theory.
Of course, permutation symmetries have been a key component of ML, both in built-in layers of NNs (q.v.~e.g., \cite{GBA}) as well as establishing algorithms in detecting them (q.v.~e.g., \cite{HW}).
Our motivations here, are different, and will focus on the intrinsic patterns in combinatorial problems.

\paragraph{Graph Properties: } 
While the initial motivation of \cite{He:2020fdg} is to study the discrete generalization of Calabi-Yau manifolds by considering the spectrum of the graph analogue of the Laplacian, much of the preparatory work is of general interest.
The Wolfram database of connected, simple, undirected graphs \cite{wolfram} was downloaded up to 100 vertices, a total of around 8000.
Such objects have a standard representation in terms of the adjacency matrix $a_{ij}$ (whose $(i,j)$-th entry corresponds to an arrow from vertex $i$ to vertex $j$) . Because we are only dealing with simple undirected graphs  here (no multi-arrows, no self-loops and only edges rather than directed arrows), $a_{ij}$ is binary, symmetric and with diagonal entry 0.
Furthermore, the matrices are not block-diagonalizable because the graphs are all connected.

There is a host of interesting properties of graphs - such as whether it is planar, what is its genus, etc. -  which have been studied over the centuries since Euler -- this is why the Wolfram database exists.
Thus, we have yet another family of labeled data exemplified by the following:
\begin{equation}
genus(\begin{array}{c}  \includegraphics[trim=12mm 0mm 0mm 0mm, clip, width=1.5in]{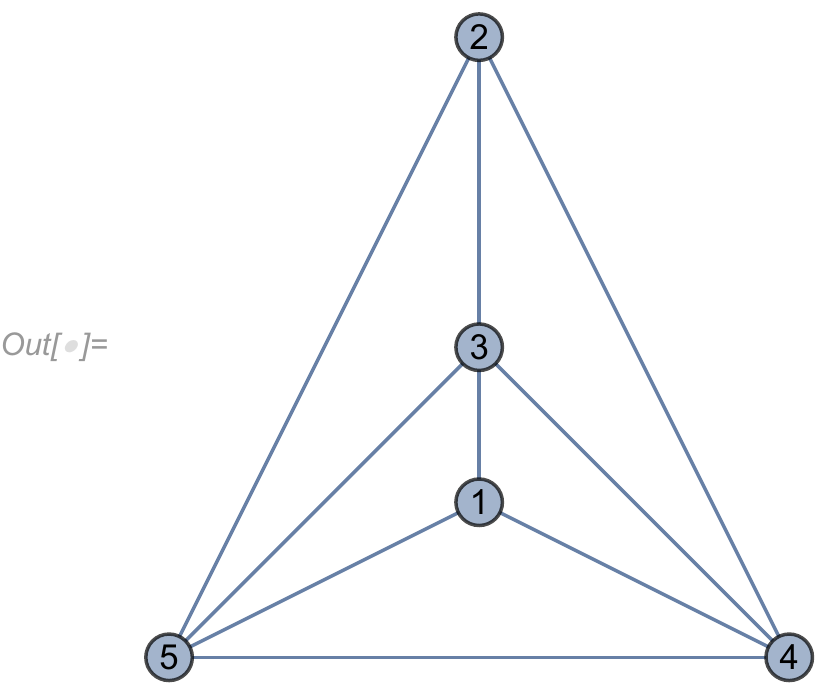}
\end{array}) = 0 \qquad \leadsto \qquad 
\left(
{\arraycolsep=1pt\def\arraystretch{0.3}
\begin{array}{ccccc}
 0 & 0 & 1 & 1 & 1 \\
 0 & 0 & 1 & 1 & 1 \\
 1 & 1 & 0 & 1 & 1 \\
 1 & 1 & 1 & 0 & 1 \\
 1 & 1 & 1 & 1 & 0 \\
\end{array}
}
\right) \longrightarrow 0 \ .
\end{equation}
Again, we can enhance the data by including random permutations of rows/columns (note that unlike Cayley tables, the rows and columns, which index the vertices, must be simultaneously permuted).
This, together with balancing, gives us various labeled data-sets of the form $\{ a_{ij} \to P \}$ with relevant property $P$, and of size $\sim 10^5$.
In particular, \cite{He:2020fdg} finds, in approximate decreasing order of accuracy: 
\begin{description}

\item[Girth: ] the min over the lengths of all cycles of the graph; it is $\infty$ if the graph is acyclic (has no cycles).
To test whether it is acylic or not, as a binary classification problem, a decision tree can get accuracy $\sim(0.95, 0.91)$.
On the other hand, a 3-category classification (of whether the girth is 3, 4, or $>4$), achieves $\sim (0.77, 0.66)$.
This is interesting since the decision of whether a graph is acyclic is easy: there is a polynomial time algorithm.

\item[Genus: ] the genus of the Riemann surface onto which the graph can be embedded. This gives a 3-way classification of $g=0$, $g=1$ and $g>1$ in analogy to Riemann uniformization for surfaces (complex curves).
Logistic regression gives accuracy $\sim (0.81, 0.72)$;

\item[Planarity: ] whether the graph can be embedded into a plane in that it
can be drawn so that no edges cross except meeting at the nodes.
This is a binary classification which logisitic regression can find accuracy $\sim (0.81, 0.62)$;

\item[Euler/Hamilton: ] if a cycle traverses all edges exactly once, it is an Euler cycle. On the other hand, if a cycle traverses all edges exactly once, it is a Hamilton cycle.  The presence of an Euler cycle is the famous K\"onigsberg bridge problem and that of a Hamilton cycle, the celebrated traveling salesman problem.  The former is known to have a polynomial time algorithm whilst the latter, NP hard.
Curiously, the binary classification problem of whether a graph has an Euler cycle or not has, with a random forest classifier, accuracy $\sim (0.73, 0.47)$, while for the the presence of a Hamilton cycle, accuracy $\sim(0.78, 0.56)$, which is comparable. Though it seems counter-intuitive that a ``hard'' and an ``easy'' problem should behave similarly to an ML algorithm, one should bear in mind that heuristic and approximate solutions to the Hamilton cycle problem abound.
Thus, stochatically, these two problems are on the same level of difficulty, in accordance with our ML results.
\end{description}

More sophisticated properties of graphs have also been explored, such as categorizing chromatic number, graph Laplacians, Ricci-flatness, etc. \cite{He:2020fdg}.
For directed graphs and associated representations of quivers, there has been a host of recent activity, especially in the context of cluster algebras. In physics, for instance, cluster mutation is identified as Seiberg duality for supersymmetric QFTs.
A systematic study was done in \cite{Bao:2020nbi} (q.v.~summary table on p7) to see how various ML algorithms detect  quiver properties such as mutation type and equivalence.

\subsection{Number Theory}\label{s:arith}
As one might intuitively suspect, number theory problems will be hard; finding simple new patterns in the primes, for example, would have unfathomable repercussions.

\paragraph{Warm-up: } 
Let us begin with primes as a warm-up.
Suppose we have a sequence of labeled data
\begin{equation}\label{primeML}
\begin{array}{l}
\{2\} \to 3 \ ; \\
\{2, 3\} \to 5 \ ; \\
\{2, 3, 5\} \to 7 \ ; \\
\{2, 3, 5, 7\} \to 11 \ ; \ldots
\end{array}
\end{equation}
One can easily check that even with millions of training data, one would be hard pressed to find an ML algorithm in predicting the next prime and that we are better off with a simple regression against the $n\log(n)$ curve of PNT.
In case the reader is worried about the large variation in the output, let us re-cast this into a binary classification problem.

We set up the data as follows:
\begin{enumerate}
\item Let $\delta(n) = 0$ or $1$ be the prime characteristic function so that it is 1 when an odd number $n$ is prime and 1 otherwise (there is no need to include even numbers);
\item Consider a ``sliding window'' of size, say, 100, and consider the list of vectors\\ $\{\delta(2i+1), \ldots, \delta(2(i+100)+1) \}_{i = 1, \ldots, 50000}$ in $\IR^{100}$. This is the list of our input;
\item For output, consider the prime characteristic of some distinct number from each windows, say, 
$\delta(2(i+100+k)+1)$ for $k=10000$.
\end{enumerate}
We thus have a binary classification problem of binary vectors, of the form ( we have ordered the set with a subscript for reference)
\begin{equation}\label{primeML2}
\begin{array}{l}
\{1, 1, 1, 0, 1, 1, \ldots, 1, 0, 1, 1, 0, 0 \}_1 \to 0 \ ; \\
\{1, 1, 0, 1, 1, 0,\ldots, 0, 1, 1, 0, 0, 0 \}_2 \to 0 \ ; \\
\ldots\\
\{1, 0, 0, 0, 0, 0,\ldots, 0, 0, 0, 0, 1, 0\}_{600} \to 1 \ ; \ldots
\end{array}
\end{equation}
Now, the primes are increasingly rare by PNT, so we down-sample the 0-hits to around  9000 each of 0 and 1.
Applying various classifiers it was found that the k-nearest neighbour (using $k=50$ and Hamming distance) worked best, at accuracy around  $(0.77, 0.60)$.

On the other hand, if we used the Liouville $\lambda$-function -  which is 1 if the number of prime factors of $n$ is even and $-1$ if odd - instead of the prime-characteristic $\delta$, we find accuracy around $(0.50, 0.001)$ with any standard ML algorithm, which is a good as randomly guessing \footnote{\label{ft:modp2}
To give another example of how difficult ``divisibility'' is, let us reconsider Footnote \ref{ft:modp}.
As mentioned, instead of fixing a prime $p$ and consider the residue of $n$ (expressed as a string of its binary digits) mod $p$, which is essentially a linear problem and can be quickly learnt by an SVM, if we did not fix $p$, and had the input as $(n,p)$ both expressed as in binary digits, then it is much more difficult to a find classifier which works.
We remark that trying the same for the even/odd property of the digits of $\pi$, say, also gives no better than random guess.
}.
This means that it is extremely difficult to predict the precise behaviour of $\lambda(n)$, as is well known.
Going back to $\delta$, it is indeed curious that we are doing quite a bit better than random guessing.
Now, it has recently come to be known \cite{AKS} that PRIMES, the problem of deciding whether $\lambda(n) = 0$ or 1, is actually polynomial time, so this is an intrinsically ``easier'' problem.
Experience tell us that a data-structure like \eqref{primeML2}, had it come from algebraic geometry over $\IC$, would be getting much higher accuracies.

\paragraph{Arithmetic Geometry: } 
Having prepared ourselves with traditional problems involving primes, it is natural to consider problems which lie between geometry and number theory, which have spear-headed much of the modern approach to arithmetic.
Initial exploration \cite{Alessandretti:2019jbs} to BSD \cite{bsd} using standard ML methods as well as topological data analysis (persistence diagrams) showed that elliptic curve data behaved not much better than frontal attacks to primes.
However, the representation used for the elliptic curve was the Weierstra\ss\ coefficients, which, like \eqref{primeML}, had huge variation in input/output structure. Indeed, as emphasized before, one should {\it normalize} the data to avoid unnecessarily large numerical range, such as the \eqref{primeML2}.  Similarly, for the Hodge number problem \eqref{cicyML} in geometry, it was natural to use $h^{1,1}$ which is a 19-channel classification problem, rather than $h^{2,1}$, which have a range in the hundreds.

With this consideration, a much more conducive representation for the elliptic curves was used \cite{He:2020tkg}: the non-trivial coefficient of the L-function. Recall that for an elliptic curve $E$, the L-function is
$\exp \left ( \sum\limits_{k=1}^\infty \frac{\#E\left(\mathbb{F}_{p^k}\right)T^k}{k} \right ) :=  \frac{L_p(X,T)}{(1-T)(1-pT)}$
and $L_p(E,T)=1 - a_p T + p T^2$ with $a_p = p+1-\#E\left(\mathbb{F}_{p}\right)$.
Thus we have labeled data-sets (around size $10^5$) \cite{lmfdb} of the form
\begin{equation}\label{apE}
(a_{p_1},\dots,a_{p_N}) \longrightarrow \mbox{ Property of } E
\end{equation}
where $p_1, \ldots, p_N$ are the first $N$ primes.
Now, with this representation, even at $N$ as small as 100, we can classify rank, torsion order, existence of integer points, etc, to accuracy  $(0.98, 0.96)$, $(1.0, 1.0)$, $(1.0,1.0)$, respectively, using a na\"{\i}ve Bayesian classifier.
Interestingly, the most difficult quantity of BSD, the Tate-Shafarevich group, obtained the least accuracy, with precision  $<0.6$. Similar results were  obtained for genus 2 curves.
In fact, other refined properties for arithmetic curves, such as those pertaining to the Sato-Tate Conjecture can also be classified by establishing data-sets as \eqref{apE}, and high accuracy can be attained \cite{He:2020kzg}.

Along the same vein, \cite{He:2020qlg} studied properties of number fields.
The type of Galois group and the order of the unit (class group size) can be predicted from the coefficients of the minimal polynomial or from that of the Dedekind zeta function, with accuracy $\gtrapprox (0.97, 0.93)$.
Likewise, the degree of the Galois extension of a dessin d'enfant can be predicted from looking at the permutation triple information of the dessin \cite{He:2020eva}.
This is quite comforting and surprising since computing Belyi maps from dessins is notoriously difficult.

In sum, we have taken problems from some of the central themes of modern number theory: BSD, the Langlands Programme and Grothendieck's Esquisse. Interestingly, the data therein possess structure which are amenable to ML, much more than classical analytical number theory (even simple problems like remainders on division, as discussed in Footnote \ref{ft:modp2}, let alone Liouville $\lambda$). It is as if arithmetic geometry is closer to geometry than to arithmetic.

\section{Conclusions and Outlook}
We have taken a casual promenade in the vast landscape of mathematics, armed purposefully only with a small arsenal of techniques from ML, in order to explore the structure of different branches, exemplified by concrete data that had been carefully compiled over the decades.
The methods employed, from SVMs to Bayesian classifiers, from simple feed-forwards NNs to decision trees, have no idea of
the intricacies of the underlying mathematics, yet they are guessing correct answers to high accuracy, sometimes even to 100\%.

This paradigm is clearly {\it useful} in at least two respects.
First, in computations which would traditionally  be too expensive and one wants a quick estimate, this ML approach would be orders of magnitude faster. For example, computing cohomology groups for algebraic varieties requires putting everything into Gr\"obner bases, which is exponentially prohibitive, but the ML, exemplified by the CICYs, only takes matter of seconds to minutes. 
Second, in the cases where accuracy consistently reaches 1.00, then one has a potential conjecture.
Of course, as with the first case, one needs to be careful about ``interpolation'' versus ``extrapolation'': we need to ensure that the ML's learning is not merely restricted the the data-set even when cross-validation is performed, but it truly has the ability to go beyond the features.  For instance, one could train on bundles of lower degree and validate on those of higher degree, or one could train on smaller graphs and validate on larger graphs, and if the accuracy remains 1.00, then one could proceed to conjectures \footnote{
One should in mind that in parallel to the traditional conjecture formulation from data, such as PNT or BSD, there are increasing number of important statistical statements in mathematics, such as distributions of ranks of elliptic curves or in prime progressions.
On the other hand, it goes without saying that one should always be careful with conjecturing formulation based on data mining: the famous Skewes number immediately springs to mind as a caveat.
}.

Of course, interpretable ML is a burgeoning field and NNs, especially, due to the complex inter-connectivities, are notoriously difficult to untangle.
This ``intelligible intelligence'' \cite{Udrescu:2019mnk} in uncovering laws of science \cite{scinet}, information geometry \cite{BN}, and symbolic mathematics \cite{DLsymb} using NNs, are becoming increasingly relevant.
In the above explorations, we have already seen exact cohomology formulae \cite{Brodie:2019dfx} and conjectured existence of hyper-plane separating simple and non-simple finite groups \cite{He:2019nzx}, etc.
Indeed, if the predictions of \cite{buzzard,ICM18} are true, then machine aided conjectures and proofs will go hand in hand within a decade.
In our sense, finding interpretable results would be extracting ``semantics'' from ``syntax'' \cite{Zilber}, from ``top-down'' to `` bottom-up''.

One might even lean toward the other extreme and forgo interpretability in certain situations.
After all, if an ML algorithm - without an analytic interpretation - does produce the correct result 100\% of the time, it is as good as an analytic formula.
On the contrary, there are many exact formulae which are ineffective.
For example, a simple consequence of Wilson's theorem, 
is that the $n$-th prime is $\left\lfloor {\frac {n!{\bmod {(}}n+1)}{n}}\right\rfloor (n-1)+2$.
The $n!$ clearly compels people not to use this when finding the $n$-th prime.
Similarly, exact expressions for cohomology over algebraic varieties do exist, to which we alluded in \S\ref{s:geo}.
Even for projective space, Bott-Borel-Weil gives
$h^q(\IC\IP^n, (\wedge^p T\IC\IP^n) \otimes \cO(k)) =
	\left\{\begin{array}{lll}
	{k+n+p+1 \choose p}{k+n \choose n-p} & q = 0 & k>-p-1,\\
	1 & q=n-p & k=-n-1,\\
	{-k-p-1 \choose -k-n-1}{-k-n-2 \choose p} & q = n & k<-n-p-1,\\
	0 & {\rm otherwise} &
	\end{array}\right.$, which is a non-trivial expression.
Take the example of the cohomology of a single line bundle of bi-degree $(-k,m)$ on a bi-degree $(2,4)$ hyper-surface in $\IC\IP^1 \times \IC\IP^3$ (which is a CICY threefold), this is known, from painful long-exact-sequence chasing, to be
\begin{equation}
h^{q}(X, \mathcal{O}_{X}(-k,m))=
	\left\{\begin{array}[c]{ll}
	(k+1)\binom{m}{3}-(k-1)\binom{m+3}{3} & q=0\quad k<\frac{(1+2m)(6+m+m^2)}{3(2+3m(1-m))}\\
	(k-1)\binom{m+3}{3}- (k+1)\binom{m}{3} & q=1\quad k>\frac{(1+2m)(6+m+m^2)}{3(2+3m(1-m))}\\
	0 & {\rm otherwise}
	\end{array}
	\right. \ . 
\end{equation}	
One can only imagine how much more complicated the expression would be for non-complete-intersection and more complicated bundles than a single line bundle!
The precise answers and region of validity are more suited for a computer programme than for any human comprehension beyond the guarantee that an exact sequence calculation would produce the right result. 
The point is that such expressions {\it in principle} exist and the principle is important. Whether they are written explicitly, or as a list of parameters and architecture of an NN, is not more enlightening either way.
While we remain agnostic, we hope the reader can appreciate both the necessity and the sometime dispensableness of interpretability.

Utility aside, our paradigm is also an approach toward understanding the fundamental {\it structure} of mathematics.
Modeling our standard ML algorithms as the ``bag of tricks'' of the working mathematician, and the various data as representing the field whence they come, we have gone through a plethora of problems ranging from geometry to arithmetic.
The collection of algorithms {\it has no idea} about the AKS algorithm, nor cohomology theory, nor graph theory, nor abstract algebra, nor arithmetic geometry \ldots, but they are seemingly picking up ``harder'' versus ``easier'' problems.
Guessing the Liouville $\lambda$ function seems to be impossible for any of the standard methods, while guessing ranks of cohomology groups of complex algebraic variety seems easy for several different classifiers and regressors.

We are tempted to approximately rank this level of difficulty - being a well aware that different disciplines are certainly intertwined and separation by sub-field is often not possible.
The ``difficulty'', we note, is {\it not} necessarily ``computational complexity''. It is correlated to it, in the several examples we have seen, exemplified by the polynomial algorithms for PRIMES, or by the stochastic search for Hamiltonian cycles in graphs, etc.
From the many experiments, we seem to have (where $<$ means less amenable to algorithmic analysis):
\begin{equation}
\begin{split}
\left[\mbox{numerical analysis}\right] < 
\left[\mbox{algebraic geometry over $\IC$ $\sim$ arithmetic geometry} \right] <  \\
\left[\mbox{algebra/representation theory} \right] < 
\left[\mbox{combinatorics} \right] < 
\left[\mbox{analytic number theory} \right]
\end{split}
\end{equation}

This ``hierarchy'' of different branches of mathematics is reminiscent of the multitude of problems, ranging from efficient numerical methods which proliferate in all areas, in contrast to the undecidability of Diophantine systems (Hilbert 10th) or to finding new patterns in the zeros of the Riemann zeta function.
One could take this to a fundamental level \cite{Zilber} with model theoretical considerations.
In categoricity theory, the theorems of  Morley-Shelah and Hart-Hrushovski-Laskowski \cite{categoricity} give a classification of number of isomorphism classes of models in various cardinalities, whereby giving a sense of the difficulty of the theory to which the problem belongs.
These and endless further explorations we shall leave to the readers' pleasure.
In the mean time, they are encouraged to submit to the topical collection \cite{TC} and to the upcoming journal {\it Data Science in the Mathematical Sciences} \cite{DSMS}.

~\\
~\\


\begin{spacing}{0.6}

\end{spacing}


\begin{thebibliography}{99}
{\small


\bibitem[AlexNet]{Alex}
A.~Krizhevsky, I.~Sutskever, G.~Hinton, ``ImageNet classification with deep convolutional neural network'', Comm.~ACM. 60 (6): 84 - 90 (2012).

\bibitem[ABH]{Alessandretti:2019jbs}
L.~Alessandretti, A.~Baronchelli and Y.~H.~He,
``Machine Learning meets Number Theory: The Data Science of Birch-Swinnerton-Dyer,''
[arXiv:1911.02008 [math.NT]].

\bibitem[AKS]{AKS}
M.~Agrawal, N.~Kayal, N.~Saxena, ``PRIMES is in P'', Annals of Mathematics. 160 (2): 781 - 793, 2002.


\bibitem[ACHN]{Altman:2018zlc} 
  R.~Altman, J.~Carifio, J.~Halverson and B.~D.~Nelson,
  ``Estimating Calabi-Yau Hypersurface and Triangulation Counts with Equation Learners,''
  JHEP {\bf 1903}, 186 (2019)
  [arXiv:1811.06490 [hep-th]].

\bibitem[AGGKRR]{Anderson:2020hux}
L.~B.~Anderson, M.~Gerdes, J.~Gray, S.~Krippendorf, N.~Raghuram and F.~Ruehle,
``Moduli-dependent Calabi-Yau and SU(3)-structure metrics from Machine Learning,''
[arXiv:2012.04656 [hep-th]].


\bibitem[AHK]{AHK}
K.~Appel, W.~Haken, ``Every Planar Map is Four Colorable. I. Discharging'', Illin.~J.~of Maths, 21 (3): 429 -- 490;
K.~Appel, W.~Haken, J.~Koch, ``Every Planar Map is Four Colorable. II. Reducibilit'', pp491–567, (1977)

\bibitem[AHO]{Ashmore:2019wzb}
A.~Ashmore, Y.~H.~He and B.~A.~Ovrut,
``Machine learning Calabi-Yau metrics,''
Fortsch. Phys. \textbf{68} (2020) no.9, 2000068
[arXiv:1910.08605 [hep-th]].

\bibitem[BB]{BB}
V.~Batyrev, L.~Borisov, ``Mirror duality and string theoretic Hodge numbers''. Inv.~Math.~126 (1): 183-203 (1996).

\bibitem[BCDL]{Brodie:2019dfx}
C.~R.~Brodie, A.~Constantin, R.~Deen and A.~Lukas,
``Machine Learning Line Bundle Cohomology,''
Fortsch. Phys. \textbf{68} (2020) no.1, 1900087
[arXiv:1906.08730 [hep-th]].

\bibitem[BFHHMX]{Bao:2020nbi}
J.~Bao, S.~Franco, Y.~H.~He, E.~Hirst, G.~Musiker and Y.~Xiao,
``Quiver Mutations, Seiberg Duality and Machine Learning,''
Phys. Rev. D \textbf{102} (2020) no.8, 086013
[arXiv:2006.10783 [hep-th]].

\bibitem[BHJM]{Bull:2018uow}
K.~Bull, Y.~H.~He, V.~Jejjala and C.~Mishra,
``Machine Learning CICY Threefolds,''
Phys. Lett. B \textbf{785} (2018), 65-72
[arXiv:1806.03121 [hep-th]].\\
--, ``Getting CICY High,''
PLB \textbf{795} (2019), 700-706
[arXiv:1903.03113 [hep-th]].

\bibitem[BN]{BN}
F.~Barbaresco, F.~Nielsen, Ed, {\it Geometric Structures of Statistical Physics, Informational Geometry and Learning,}
SPIGL'20 proceedings, Les Houches, Springer 2021.\\
F.~Nielsen, Ed, , {\it Progress in Information Geometry: Theory and Applications},
Springer 2021

\bibitem[BSD]{bsd}
B.~Birch, P.~Swinnerton-Dyer, ``Notes on Elliptic Curves (II)'', J.~Reine Angew.~Math. 165 (218): 79 - 108 (1965);
experiments on EDSAC-2, Cambridge.

\bibitem[Buz]{buzzard}
K.~Buzzard, ``The future of Mathematics,'' \url{https://wwwf.imperial.ac.uk/~buzzard/one_off_lectures/msr.pdf};
\url{https://www.youtube.com/watch?v=Dp-mQ3HxgDE}

\bibitem[Cat]{categoricity}
S.~Shelah, ``Classification theory and the number of nonisomorphic models'', 
Studies in Logic and the Found.~of Maths, vol. 92, IX, 1.19, p.49 (1990).\\
B.~Hart, E.~Hrushovski, M.~Laskowski, ``The Uncountable Spectra of Countable Theories'', Annals of Maths. 152 (1): 
207257. arXiv:math/0007199\\
B.~Zilber, ``Categoricity'', AMS G\"odel Lecture, 2003, \url{https://people.maths.ox.ac.uk/zilber/godel.pdf}

\bibitem[CCHKLN]{Carifio:2017nyb}
J.~Carifio, W.~J.~Cunningham, J.~Halverson, D.~Krioukov, C.~Long and B.~D.~Nelson,
``Vacuum Selection from Cosmology on Networks of String Geometries,''
Phys. Rev. Lett. \textbf{121} (2018) no.10, 101602
[arXiv:1711.06685 [hep-th]].

\bibitem[CDLS]{cicy}
P.~Candelas, A.~M.~Dale, C.~A.~Lutken, R.~Schimmrigk,
  ``Complete Intersection Calabi-Yau Manifolds,''
  Nucl.\ Phys.\ B {\bf 298}, 493 (1988).\\
M.~Gagnon, Q.~Ho-Kim,
  ``An Exhaustive list of complete intersection Calabi-Yau manifolds,''
  Mod.\ Phys.\ Lett.\ A {\bf 9} (1994) 2235.
\\
T.~Hubsch,
{\it Calabi-Yau manifolds: A Bestiary for physicists},
World Scientific, 1994, ISBN 9810206623  

\bibitem[3CinG]{CGCR}
T.~Coates, M.~Gross, A.~Corti, M.~Reid, ``Classification, Computation, and Construction: New Methods in Geometry,''
\url{http://geometry.ma.ic.ac.uk/3CinG/}

\bibitem[Chu]{church}
A.~Church, ``An Unsolvable Problem of Elementary Number Theory''. Amer.~ J.~of Math. 58 (2): 345 - 363  (1936).

 \bibitem[CHKN]{Carifio:2017bov}
 J.~Carifio, J.~Halverson, D.~Krioukov and B.~D.~Nelson,
  ``Machine Learning in the String Landscape,''
  JHEP {\bf 1709}, 157 (2017)
  [arXiv:1707.00655 [hep-th]].

\bibitem[CHLZ]{Chen:2020dxg}
H.~Y.~Chen, Y.~H.~He, S.~Lal and M.~Z.~Zaz,
``Machine Learning Etudes in Conformal Field Theories,''
[arXiv:2006.16114 [hep-th]].

\bibitem[CHLM]{Chen:2020jjw}
H.~Y.~Chen, Y.~H.~He, S.~Lal and S.~Majumder,
``Machine Learning Lie Structures \& Applications to Physics,''
[arXiv:2011.00871 [hep-th]].

\bibitem[CJKP]{Craven:2020bdz}
V.~Jejjala, A.~Kar and O.~Parrikar,
``Deep Learning the Hyperbolic Volume of a Knot,''
Phys. Lett. B \textbf{799} (2019), 135033
[arXiv:1902.05547 [hep-th]].\\
J.~Craven, V.~Jejjala and A.~Kar,
``Disentangling a Deep Learned Volume Formula,''
[arXiv:2012.03955 [hep-th]].


\bibitem[CL]{Constantin:2018hvl}
A.~Constantin and A.~Lukas,
``Formulae for Line Bundle Cohomology on Calabi-Yau Threefolds,''
Fortsch. Phys. \textbf{67} (2019) no.12, 1900084
[arXiv:1808.09992 [hep-th]].

\bibitem[Coq]{Coq}
The Coq Proof Assistant, \url{https://coq.inria.fr/}

\bibitem[Coqu]{Coquand}
T.~Coquand, ``An analysis of Girard’s paradox,'' Proc.~IEEE Symposium on Logic in Computer Science, 227 - 236 (1986).


\bibitem[DHLL]{Deen:2020dlf}
R.~Deen, Y.~H.~He, S.~J.~Lee and A.~Lukas,
``Machine Learning String Standard Models,''
[arXiv:2003.13339 [hep-th]].

\bibitem[DSMS]{DSMS}
Y.-H.~He, Manag.~Ed., {\it Data Science in the Mathematical Sciences,} World Scientific, 2021, to appear.

\bibitem[DLQ]{Douglas:2020hpv}
M.~R.~Douglas, S.~Lakshminarasimhan and Y.~Qi,
``Numerical Calabi-Yau metrics from holomorphic networks,''
[arXiv:2012.04797 [hep-th]].


\bibitem[EF]{Erbin:2020srm}
H.~Erbin and R.~Finotello,
``Inception Neural Network for Complete Intersection Calabi-Yau 3-folds,''
[arXiv:2007.13379 [hep-th]].\\
--``Machine learning for complete intersection Calabi-Yau manifolds: a methodological study,''
[arXiv:2007.15706 [hep-th]].

\bibitem[Fre]{Frege}
G.~Frege, {\it Die Grundlagen der Arithmetik. Eine logisch-mathematische Untersuchung über den Begriff der Zahl},
Verlag von Wilhelm Koebner (1884).


\bibitem[GAP]{gap}
  The GAP~Group, \emph{GAP -- Groups, Algorithms, and Programming, 
  Version 4.9.2};  2018,
  \url{https://www.gap-system.org}
  
\bibitem[GBA]{GBA}
Ian Goodfellow, Yoshua Bengio, Aaron Courville,
{\it Deep Learning}, ISBN: 9780262035613, MIT Press, 2016.
  
\bibitem[GBML]{GBML}
J.~De Loera, S.~Petrovic, L.~Silverstein, D.~Stasi, D.~Wilburne. 
``Random monomial ideals,'' J.~Algebra 519, 440 - 473, 2019.\\
D.~Peifer, M.~Stillman, D.~Halpern-Leistner,
``Learning selection strategies in Buchberger's algorithm'',
[arXiv:2005.01917]

\bibitem[GHRS]{Gukov:2020qaj}
S.~Gukov, J.~Halverson, F.~Ruehle and P.~Su\l{}kowski,
``Learning to Unknot,''
[arXiv:2010.16263 [math.GT]].

\bibitem[God]{Godel}
K.~G\"odel, ``\"Uber formal unentscheidbare S\"atze der Principia Mathematica und verwandter Systeme, I.'' 
Monatshefte f\"ur Mathematik und Physik 38: 173-198 (1931).

\bibitem[Gon]{Gonthier}
G.~Gonthier et al., 
``Formal Proof—The Four- Color Theorem'' Not.~AMS, 2005
\\
--, ``A Machine-Checked Proof of the Odd Order Theorem'', 
Interactive Theorem Proving pp 163 -- 179,  Lect. Notes in CS, Vol 7998 (2013)


\bibitem[GRdB]{grdb}
The Graded Ring Database,
\url{http://www.grdb.co.uk/}
\\
The $C^3$NG collaboration: \url{http://geometry.ma.ic.ac.uk/3CinG/index.php/team-members-and-collaborators/}
Data at:
\url{http://geometry.ma.ic.ac.uk/3CinG/index.php/data/}
\url{http://coates.ma.ic.ac.uk/fanosearch/}

\bibitem[HeCY]{He:2018jtw}
Y.-H.~He,
``The Calabi-Yau Landscape: from Geometry, to Physics, to Machine-Learning,''
[arXiv:1812.02893 [hep-th]]. To appear, Springer.

\bibitem[HeEnc]{He:2020bfv}
Y.~H.~He,
``Calabi-Yau Spaces in the String Landscape,''
Entry to Oxford Res.~Encyclo.~of Physics, B.~Foster Ed., OUP, 2020
[arXiv:2006.16623 [hep-th]] 

\bibitem[HeTalk]{HeTalks}
q.v., some of the author's talks at StringData 2020 \url{https://www.youtube.com/watch?v=GqoqxFsaogY};
Clifford 2020 \url{http://wlkt.ustc.edu.cn/video/detail_5372_24779.htm}; and CMSA, Harvard
\url{https://www.youtube.com/watch?v=zj_Xc2QG-vw}

\bibitem[HeDL]{He:2017aed} 
  Y.~H.~He,
  ``Deep-Learning the Landscape,''
  arXiv:1706.02714 [hep-th].
   {\it Science}, vol 365, issue 6452, Aug 2019.
\\
--,
``Machine-learning the string landscape,''
  Phys.\ Lett.\ B {\bf 774}, 564 (2017).
  
\bibitem[HDKL]{TC}
Y-H.~He, P.~Dechant, A.~Kaspryzyk, A.~Lukas, Ed. ``Machine-learning mathematical structures,''
topical collection for Advances in Applied Clifford Algebras, Birkh\"auser, Springer, call open:
\url{https://www.springer.com/journal/6/updates/18581430}

\bibitem[HHP]{He:2020eva}
Y.~H.~He, E.~Hirst and T.~Peterken,
``Machine-Learning Dessins d'Enfants: Explorations via Modular and Seiberg-Witten Curves,''
[arXiv:2004.05218 [hep-th]].


\bibitem[HeLee]{He:2019vsj}
Y.~H.~He and S.~J.~Lee,
``Distinguishing elliptic fibrations with AI,''
Phys. Lett. B \textbf{798} (2019), 134889
[arXiv:1904.08530 [hep-th]].

\bibitem[HK]{He:2019nzx}
Y.~H.~He and M.~Kim,
``Learning Algebraic Structures: Preliminary Investigations,''
[arXiv:1905.02263 [cs.LG]].



\bibitem[HL]{He:2020lbz}
Y.~H.~He and A.~Lukas,
``Machine Learning Calabi-Yau Four-folds,''  to appear PLB,
[arXiv:2009.02544 [hep-th]].

\bibitem[HLOac]{He:2020tkg}
Y.~H.~He, K.~H.~Lee and T.~Oliver,
``Machine-Learning Arithmetic Curves,''
[arXiv:2012.04084 [math.NT]].

\bibitem[HLCnf]{He:2020qlg}
Y.~H.~He, K.~H.~Lee and T.~Oliver,
``Machine-Learning Number Fields,''
[arXiv:2011.08958 [math.NT]].


\bibitem[HLOst]{He:2020kzg}
Y.~H.~He, K.~H.~Lee and T.~Oliver,
``Machine-Learning the Sato--Tate Conjecture,''
[arXiv:2010.01213 [math.NT]].



  
\bibitem[HM]{He:2015yoa}
Y.~H.~He and J.~McKay,
``Sporadic and Exceptional,''
[arXiv:1505.06742 [math.AG]].

  
\bibitem[HeUni]{He:2020mgx}
Y.-H.~He,
``Universes as Big Data,''
[arXiv:2011.14442 [hep-th]], to appear IJMPA.

\bibitem[HeYau]{He:2020fdg}
Y.~H.~He and S.~T.~Yau,
``Graph Laplacians, Riemannian Manifolds and their Machine-Learning,''
[arXiv:2006.16619 [math.CO]].


\bibitem[HW]{HW}
D.~Helmbold, M.~Warmuth,``Learning Permutations with exponential weights'', J.~Machine Learning Res.~2009 (10) 1705-1736.

\bibitem[ICM18]{ICM18}
J.~Davenport, B.~Poonen, J.~Maynard, H.~Helfgott, P.~H.~Tiep, L.~Cruz-Filipe,
``Machine-Assisted Proofs'', 

\bibitem[ICMS]{ICMS}
Mathematical Software, \url{http://icms-conference.org/}, q.v., also the author's talk at ICMS2020.

\bibitem[IMO]{IMO}
The IMO Grand Challenge,
\url{https://imo-grand-challenge.github.io/}

\bibitem[IMWRR]{scinet}
R.~Iten, T.~Metger, H.~Wilming, L.~del Rio, R.~Renner,
``Discovering Physical Concepts with Neural Networks'',
Phys. Rev. Lett. 124, 010508,  2020



\bibitem[JPM]{Jejjala:2020wcc}
V.~Jejjala, D.~K.~Mayorga Pena and C.~Mishra,
``Neural Network Approximations for Calabi-Yau Metrics,''
[arXiv:2012.15821 [hep-th]].


\bibitem[KNOT]{knots}
The Knots Atlas, \url{http://katlas.org/wiki/Main_Page}

\bibitem[KrSk]{Kreuzer:2000xy} 
  M.~Kreuzer and H.~Skarke,
  ``Complete classification of reflexive polyhedra in four-dimensions,''
  Adv.\ Theor.\ Math.\ Phys.\  {\bf 4}, 1209 (2002)
  [hep-th/0002240].
  \url{http://hep.itp.tuwien.ac.at/~kreuzer/CY/}

\bibitem[KlSch]{Klaewer:2018sfl}
D.~Klaewer, L.~Schlechter,
``Machine Learning Line Bundle Cohomologies of Hypersurfaces in Toric Varieties,''
PLB \textbf{789} (2019), 438-443
[arXiv:1809.02547 [hep-th]].

\bibitem[KS]{Krefl:2017yox}
D.~Krefl and R.~K.~Seong,
``Machine Learning of Calabi-Yau Volumes,''
Phys. Rev. D \textbf{96} (2017) no.6, 066014
[arXiv:1706.03346 [hep-th]].


\bibitem[KrSy]{Krippendorf:2020gny}
S.~Krippendorf and M.~Syvaeri,
``Detecting Symmetries with Neural Networks,''
[arXiv:2003.13679 [physics.comp-ph]].

\bibitem[Lean]{Lean}
Lean Theorem Prover, \url{https://leanprover.github.io/}

\bibitem[Leib]{Leibniz}
G.~Leibniz, {\it Opera omnia nunc primum collecta}, Dutens, 6 vols., Geneva 1768.\\
W.~Lenzen, ``Leibniz’s Logic'' in {\it Handbook of the History of Logic}, D.~M.~Gabbay \& J.~Woods (eds.), 
Elsevier (2004).

\bibitem[LieArt]{LieArt}
R. Feger, T. W. Kephart and R. J. Saskowski, ``LieART 2.0 – A Mathematica Application for Lie Algebras and Representation Theory,'' Comput. Phys. Commun. 257 (2020), 107490 [arXiv:1912.10969 [hep-th]]. 
\url{https://lieart.hepforge.org/}

\bibitem[LC]{DLsymb}
G.~Lample, F.~Charton
``Deep Learning for Symbolic Mathematics'',
	arXiv:1912.01412 [cs.SC]


\bibitem[LMFdB]{lmfdb}
The L-functions \& Modular Forms Database,
\url{http://www.lmfdb.org/}

\bibitem[LS]{Larfors:2020ugo}
M.~Larfors and R.~Schneider,
``Explore and Exploit with Heterotic Line Bundle Models,''
Fortsch. Phys. \textbf{68} (2020) no.5, 2000034
[arXiv:2003.04817 [hep-th]].


\bibitem[MAG]{magma}
Magma Comp.~Algebra System,
\url{http://magma.maths.usyd.edu.au/}

\bibitem[MHK]{Minhyong}
Minhyong Kim, Private communications.


\bibitem[M-L]{M-L}
P.~Martin-L\"of, ``An intuitionistic theory of types: predicative part, Logic Colloquium'' (Bristol, 1973), 73--118. 
Studies in Logic and the Foundations of Maths, Vol. 80, Amsterdam, 1975.


\bibitem[M2]{m2}
D.~Grayson, M.~Stillman,
``Macaulay2, a software system for research in algebraic geometry'',
Available at \url{https://faculty.math.illinois.edu/Macaulay2/}



\bibitem[New]{Newborn}
M.~Newborn, {\it Automated Theorem Proving: Theory and Practice}, Springer 2001.

\bibitem[NSS]{NSS}
A.~Newell, J.~Shaw, H.~Simon, Computer programme (1956) \& ``Report on a general problem-solving program,'' Proc.~Int.~Conf.~Information Processing, pp. 256 - 264 (1959).

\bibitem[OT]{Otsuka:2020nsk}
H.~Otsuka and K.~Takemoto,
``Deep learning and k-means clustering in heterotic string vacua with line bundles,''
JHEP \textbf{05} (2020), 047
[arXiv:2003.11880 [hep-th]].


\bibitem[QFTNN]{QFTNN}
K.~Hashimoto,
``AdS/CFT correspondence as a deep Boltzmann machine,''
Phys. Rev. D \textbf{99} (2019) no.10, 106017
[arXiv:1903.04951 [hep-th]].
\\
E.~d.~Koch, R.~de Mello Koch and L.~Cheng,
``Is Deep Learning a Renormalization Group Flow?,''
[arXiv:1906.05212 [cs.LG]].
\\
J.~Halverson, A.~Maiti and K.~Stoner,
``Neural Networks and Quantum Field Theory,''
[arXiv:2008.08601 [cs.LG]].
\\
V.~Vanchurin, ``The world as a neural network.'' Entropy 22.11 (2020): 1210.
arXiv:2008.01540.
\\
S.~Wolfram, \url{https://www.wolframphysics.org/}



\bibitem[Rue]{Ruehle:2017mzq}
F.~Ruehle,
``Evolving neural networks with genetic algorithms to study the String Landscape,''
JHEP \textbf{08} (2017), 038
[arXiv:1706.07024 [hep-th]].

\bibitem[RuePR]{Ruehle:2020jrk}
F.~Ruehle,
``Data science applications to string theory,''
Phys. Rept. \textbf{839} (2020), 1-117

\bibitem[RW]{RW}
A.~Whitehead, B.~Russell, {\it Principia mathematica}, CUP (1910).

\bibitem[SAGE]{sage}
SageMath, ``the Sage Mathematics Software System'',
   The Sage Developers, \url{http://www.sagemath.org}

\bibitem[Sze]{Szegedy}
C.~Szegedy, \url{https://scale.com/interviews/christian-szegedy}

\bibitem[Sing]{singular}
W.~Decker, G-M.~Greuel, G.~Pfister, H.~Sch{\"o}nemann,
{\sc Singular}, {A} computer algebra system for polynomial computations.
{http://www.singular.uni-kl.de}

\bibitem[TTH]{TTH}
A.~Tanaka, A.~Tomiya, K.~Hashimoto,
{\it Deep Learning and Physics}, Springer, to appear 2021.


\bibitem[Tur]{Turing}
A.~M.~Turing, ``On Computable Numbers, with an Application to the Entscheidungsproblem'', Proc.~LMS, 2, 42, pp. 230-65 (1937).

\bibitem[Voe]{Voe}
V.~Voevodsky, ``A Very Short Note on Homotopy Lambda Calculus'' (2006); 
``The Equivalence Axiom and Univalent Models of Type Theory,'' arXiv:1402.5556;
\url{https://www.math.ias.edu/vladimir/Univalent_Foundations}

\bibitem[UAT]{UAT}
G.~Cybenko, ``Approximation by superpositions of a sigmoidal function'', Math.~of Control, Signals, and Systems, 2 (4), 303 - 314, 1989;\\
K.~Hornik, ``Approximation capabilities of multilayer feedforward networks'', Neural Networks, 4(2), 251 - 257, 1991.\\
P.~Kidger, T.~Lyons, ``Universal Approximation with Deep Narrow Networks'', Conference on Learning Theory. 
[arXiv:1905.08539]\\
B.~Hanin, ``Approximating Continuous Functions by ReLU Nets of Minimal Width,''
ArXiv:1710.11278.

\bibitem[UT]{Udrescu:2019mnk}
S.~M.~Udrescu and M.~Tegmark,
``AI Feynman: a Physics-Inspired Method for Symbolic Regression,''
[arXiv:1905.11481 [physics.comp-ph]].

\bibitem[Wolf]{wolfram}
Wolfram Research, Inc.,~{\it Mathematica}, Champaign, IL  \url{www.wolfram.com}

\bibitem[Xena]{Xena}
The Xena Project,
\url{https://wwwf.imperial.ac.uk/~buzzard/xena/}

\bibitem[YGH]{Yang:2019kup}
C.~N.~Yang, M.~L.~Ge and Y.~H.~He, Ed.
``Topology and Physics,'' with contributions from Atiyah, Penrose, Witten, et al.,
WS 2019. ISBN: 978-981-3278-49-3
\url{https://doi.org/10.1142/11217}

\bibitem[Wil]{Wilson}
R.~Wilson, {\it The Finite Simple Groups}, Springer, 2009.

\bibitem[Zilb]{Zilber}
B.~Zilber, Private communications and work in progress.




}
\end{thebibliography}
\end{document}